\documentclass[10pt,conference,a4paper]{IEEEtran}
\IEEEoverridecommandlockouts
\usepackage{cite}
\usepackage{amsmath,amssymb,amsfonts}
\usepackage{graphicx}
\usepackage{textcomp}
\usepackage{xcolor}
\usepackage{url}
\usepackage{booktabs}
\usepackage{tabularx}
\usepackage{multirow}
\usepackage{rotating}
\usepackage{comment}
\begin{document}

\title{EigenTrack: Temporal Spectral Analysis of Hidden Activations for Hallucination and OOD Detection}

{\author{\IEEEauthorblockN{Davide Ettori}
\IEEEauthorblockA{\textit{University of Illinois Chicago}\\
Chicago, IL, USA\\
detto3@uic.edu}
\and
\IEEEauthorblockN{Nastaran Darabi}
\IEEEauthorblockA{\textit{University of Illinois Chicago}\\
Chicago, IL, USA}
\IEEEauthorblockA{ndarab2@uic.edu}
\hspace{2in}\and
\IEEEauthorblockN{Sina Tayebati}
\IEEEauthorblockA{\textit{University of Illinois Chicago}\\
Chicago, IL, USA}
\IEEEauthorblockA{stayeb3@uic.edu}
\and
\IEEEauthorblockN{Ranganath Krishnan}
\IEEEauthorblockA{\textit{Capital One AI Labs}\\
USA}
\IEEEauthorblockA{ranganath.krishnan@capitalone.com}
\and
\IEEEauthorblockN{Mahesh Subedar}
\IEEEauthorblockA{\textit{Intel Labs}\\
USA}
\IEEEauthorblockA{mahesh.subedar@intel.com}
\and
\IEEEauthorblockN{Omesh Tickoo}
\IEEEauthorblockA{\textit{Intel Labs}\\
USA}
\IEEEauthorblockA{omesh.tickoo@intel.com}
\and
\IEEEauthorblockN{Amit Ranjan Trivedi}
\IEEEauthorblockA{\textit{University of Illinois Chicago}\\
Chicago, IL, USA\\
amitrt@uic.edu}
}


\maketitle

\begin{abstract}
Large language and vision-language models (LLMs and VLMs) offer broad utility but remain prone to hallucination and out-of-distribution (OOD) errors. We propose \textit{EigenTrack}, an interpretable real-time detector that characterizes model dynamics through the spectral geometry of hidden activations. EigenTrack constructs sliding-window activation matrices, extracts covariance spectrum statistics including leading eigenvalues, spectral gaps, entropy, and Random Matrix Theory features based on divergence from the Marchenko-Pastur law, and streams these signals into a lightweight recurrent classifier to model temporal evolution. This design detects shifts toward noise-like representation regimes, enabling early identification of hallucination and OOD behavior before surface-level errors appear. Unlike black-box and grey-box methods that require multiple generations or log-probability access, EigenTrack operates with a single forward pass and no resampling. Unlike prior white-box approaches based on static snapshots, it preserves temporal context and aggregates global spectral signals across layers. We evaluate EigenTrack on LLaMa, Qwen, Mistral, and LLaVa models for hallucination detection on HaluEval and OOD detection using WebQuestions and Eurlex, achieving consistent AUROC gains with favorable accuracy-latency tradeoffs. Early detection enables efficient termination of failing generations, reducing computation while improving LLMs and VLMs reliability.
\end{abstract}


\section{Introduction and Prior Works}

Large language and vision-language models (LLMs and VLMs) are increasingly deployed in high-stakes domains such as healthcare, law, and finance, yet remain unreliable due to hallucination and severe degradation under out-of-distribution (OOD) inputs \cite{tayebati2025cap,jayasuriya2025sparc}. Early detection approaches relied on surface-level signals such as softmax confidence \cite{hendrycks2017baseline} or semantic-entropy surrogates \cite{farquhar2024entropy}, but ignored internal model dynamics and often failed under domain shift.

More recent methods differ by access level. \textit{Black-box} approaches, including SelfCheckGPT \cite{selfcheckgpt}, CoNLI \cite{conli}, and Cost-Effective HD \cite{costeffective}, rely on ensembles or multiple generations to capture uncertainty, at the cost of high latency. \textit{Grey-box} methods such as DetectGPT \cite{detectgpt}, Fast-DetectGPT \cite{fastdetectgpt}, and Glimpse \cite{glimpse} exploit log-probability curvature or partial logits, but remain snapshot-based and lack temporal context. \textit{White-box} detectors probe hidden states directly, but often lack generality or temporal modeling. For example, MIND \cite{su2024mind} streams activations without capturing evolution over time, LapEigvals \cite{binkowski2025lapeigvals} analyzes spectra at individual steps, ReDeEP \cite{sun2025redeep} relies on retrieval signals, and \cite{sriramanan2024attention} tracks attention shifts. Conformal prediction at the output layer is leveraged for uncertainty and lack of reliability in \cite{stutts2024mutual,stutts2023lightweight}.

We argue that spectral signatures provide a principled foundation for hallucination and OOD detection by compressing high-dimensional hidden activations into compact descriptors of representation geometry. Eigenvalue distributions, entropy, and spectral gaps are sensitive to low-rank correlations and instabilities that arise under distribution shift or hallucination. Unlike token-level probabilities, which reflect only output-layer uncertainty, spectral statistics integrate information across hidden layers and capture global uncertainty dynamics. Moreover, spectral analysis enables the use of Random Matrix Theory by comparing empirical activation spectra to the Marchenko-Pastur (MP) law \cite{marchenko1967distribution} using KL divergence and Wasserstein distance. Deviations from this noise baseline provide a compact and interpretable indicator of structural breakdown for hallucination or OOD behavior.

\begin{figure}[t!]
    \centering
    \includegraphics[width=\linewidth]{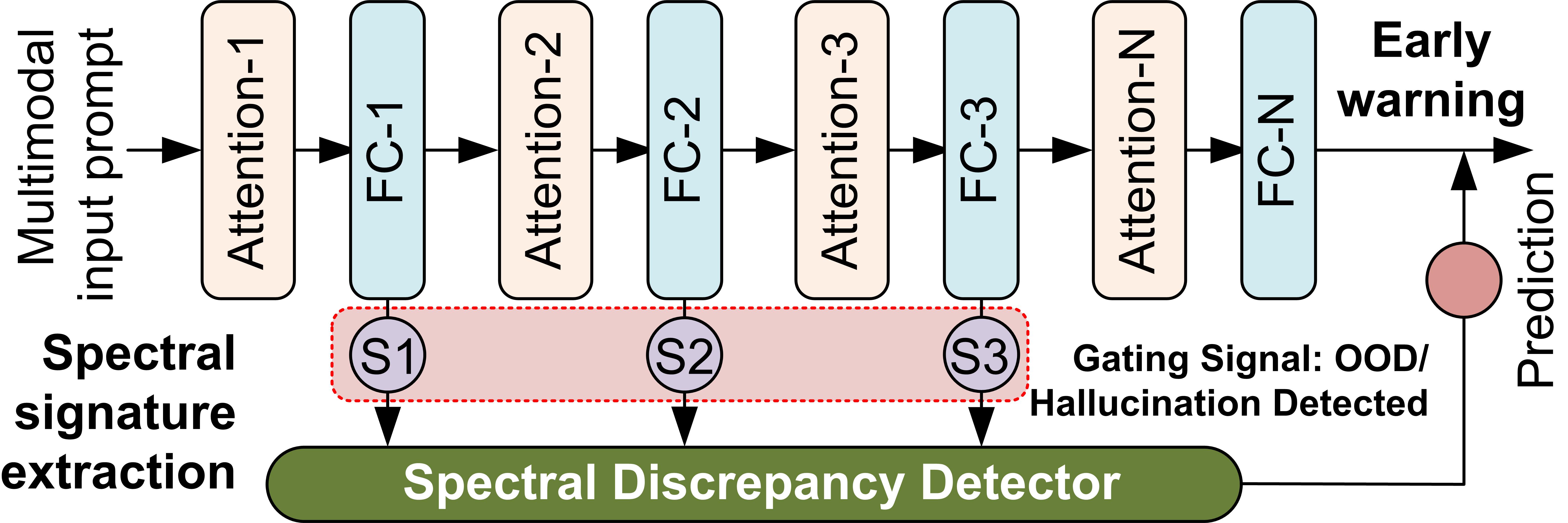}
    \caption{\textbf{EigenTrack.} Spectral signatures, including entropy, spectral gaps, and divergence from a random matrix baseline, are extracted from intermediate feed-forward layers and streamed into a recurrent spectral discrepancy detector. Tracking their temporal evolution enables early detection of failure.\vspace{-5pt}}
    \label{fig:architecture}
\end{figure}

Building on this insight, EigenTrack computes covariance spectra over a sliding window of hidden activations and streams the resulting spectral statistics into a lightweight recurrent classifier (Fig.~\ref{fig:architecture}). Prior spectral methods such as RankFeat \cite{song2022rankfeat}, SpectralGap \cite{gu2025spectralgap}, and SNoJoE \cite{mei2024snojoe} discriminate OOD from in-distribution samples using singular values or spectral gaps, but operate on static snapshots and ignore temporal evolution. In contrast, EigenTrack transforms streaming spectral features into low-dimensional temporal trajectories that reveal how uncertainty accumulates during generation. The key novelty lies in combining temporal modeling with RMT-grounded spectral features computed from global hidden activations, yielding an interpretable and effective detector for both hallucination and OOD. Our contributions are:
\begin{itemize}
    \item We introduce \textit{EigenTrack}, a real-time detector that models the temporal evolution of multi-layer spectral features to identify hallucination and OOD behavior.
    \item We propose an interpretable set of spectral indicators that link eigenvalue dynamics to latent representation shifts.
    \item We develop a data-generation pipeline for hallucination detection based on multi-LLM interaction.
\end{itemize}

EigenTrack achieves AUROC values of 0.82 to 0.94 for hallucination detection and 0.85 to 0.96 for OOD detection across LLMs and VLMs. On LLaMa-7B, it reaches 0.89 AUROC for hallucination and 0.92 for OOD, outperforming baselines such as HaloScope, LapEigvals, and SelfCheckGPT while using only a lightweight recurrent head. By enabling early termination of failing generations, EigenTrack reduces generation cost while increasing reliability.

\section{Background}

Random Matrix Theory (RMT) provides asymptotic laws for the spectra of large random matrices and serves as a principled null model for high-dimensional statistics \cite{wigner1958distribution,marchenko1967distribution}. When empirical spectra deviate from these laws, the deviation signals structure beyond isotropic noise. This property has made RMT a useful analytical tool for probing signal versus noise in high-dimensional systems. A central object in RMT is the \emph{spiked covariance model} \cite{paul2007asymptotics}, which describes data composed of a low-rank signal embedded in isotropic noise. The population covariance takes the form
\[
\boldsymbol{\Sigma} = \sigma^2 \mathbf{I}_p + \sum_{i=1}^k \theta_i u_i u_i^T,
\]
where $\theta_i$ denote spike strengths and $u_i$ are orthonormal signal directions. In this model, meaningful structure manifests as isolated spectral components that separate from the noise bulk. In this work, we apply this framework to covariance matrices constructed from LLM hidden activations, using RMT as a reference to distinguish structured representation dynamics from noise-like behavior.

\subsection{Wigner Matrices and the Semicircle Law}

A Wigner matrix is a symmetric matrix $W \in \mathbb{R}^{n \times n}$ with independent, zero-mean entries of variance $\sigma^2/n$ in the upper triangle and $W_{ij} = W_{ji}$. As $n \to \infty$, the empirical eigenvalue density converges almost surely to the Wigner semicircle law,
\[
\rho_{\mathrm{W}}(\lambda) = \frac{1}{2\pi\sigma^2}\sqrt{4\sigma^2 - \lambda^2}, \quad |\lambda| \le 2\sigma,
\]
and is zero outside this interval. This law provides a canonical noise-only baseline for symmetric random matrices and motivates the use of spectral bulk behavior as a proxy for unstructured noise.

\subsection{Sample Covariance and the Marchenko-Pastur Law}

For data matrices $X \in \mathbb{R}^{d \times N}$ with independent rows or columns and covariance $\sigma^2 I$, the sample covariance $C = \tfrac{1}{N} X X^\top$ has eigenvalues that converge to the Marchenko-Pastur (MP) distribution,
\[
\rho_{\mathrm{MP}}(\lambda) = \frac{1}{2\pi\sigma^2 q \lambda}\sqrt{(\lambda_{+}-\lambda)(\lambda-\lambda_{-})}, 
\quad \lambda_{\pm} = \sigma^2 (1 \pm \sqrt{q})^2,
\]
where $q = d/N$ is the aspect ratio. The MP law serves as a null baseline for covariance spectra generated by high-dimensional isotropic noise. To quantify deviations from this baseline, we compare the empirical spectral density $p(\lambda)$ to the MP reference $q(\lambda)$ using KL divergence,
\[
D_{\mathrm{KL}}(p\|q) = \int p(\lambda)\log\frac{p(\lambda)}{q(\lambda)} \, d\lambda,
\]
and the 1-Wasserstein distance,
\[
W_1(p,q) = \int_0^1 \big| F_p^{-1}(u) - F_q^{-1}(u) \big| \, du.
\]
KL divergence measures how unlikely the observed spectrum is under the MP baseline, while Wasserstein distance captures the average mass displacement required to transform one distribution into the other. Persistent eigenvalues outside MP support $[\lambda_{-}, \lambda_{+}]$ indicate low-rank correlations and structure.

\subsection{BBP Phase Transition}

The Baik-Ben Arous-Peche (BBP) phase transition \cite{baik2005phase} characterizes when a low-rank signal becomes detectable in high-dimensional noise. If the spike strength is below a critical threshold, the corresponding sample eigenvalues remain embedded within the MP bulk and are statistically indistinguishable from noise. Only when
\[
\theta > \sigma^2 (1 + \sqrt{c}),
\]
equivalently when population eigenvalues exceed $\lambda_{+} = \sigma^2 (1 + \sqrt{c})^2$, do sample eigenvalues separate from the bulk and emerge as isolated outliers. These detached eigenvalues carry signal, while eigenvalues inside the bulk represent noise-dominated directions. This transition provides a concrete criterion for distinguishing structured from unstructured regimes.

RMT has been widely applied to analyze deep learning systems by relating spectra of weights, activations, or Hessians to training dynamics, implicit regularization, and generalization behavior \cite{pennington2017rmt,martin2017implicit}. These results motivate the use of spectral statistics as compact probes of representation geometry in modern neural networks.

\subsection{Why These Results Apply to LLM Activations}

Modern decoder-only LLMs apply LayerNorm at every transformer block \cite{radford2019language}, centering activations and normalizing variance at each layer. Combined with the near-orthogonality induced by weight decay and Adam-style optimization \cite{kobayashi2024weight}, per-token hidden activations are approximately mean-zero and weakly correlated across dimensions. In addition, hidden layer widths are large, placing activation matrices firmly in the high-dimensional regime where RMT predictions are most accurate.

Under these conditions, the eigenvalue distribution of a sliding-window activation matrix $H_t \in \mathbb{R}^{N \times d}$ is well approximated by the Marchenko-Pastur law and serves as a principled null baseline. Our RMT-grounded hypothesis is that during anomalous behavior, including hallucination and OOD generation, representation structure weakens and activation spectra drift toward this noise baseline. In contrast, factual and in-distribution generation exhibits stronger low-rank structure, reflected by spectral spikes that exceed the BBP threshold \cite{baik2005phase}. This behavior has been observed in related spectral analyses of VLM robustness \cite{darabi2025eigenshield, ettori2025rmtkdrandommatrixtheoretic} and is confirmed empirically by EigenTrack. In our experiments, anomalous sequences move closer to the MP baseline, while factual sequences display clearer spectral separation, and the most predictive signals are precisely divergence-to-baseline and eigenvalue-based statistics (Fig.~\ref{fig:feature_sequence}, Table~\ref{tab:auroc_results}, Fig.~\ref{fig:SHAP}).

\section{EigenTrack: Methodology}

EigenTrack detects hallucination and out-of-distribution (OOD) behavior by transforming hidden activations of LLMs and VLMs into compact spectral descriptors and modeling their temporal evolution. It is designed to identify when representation structure degrades toward noise-like regimes that precede observable generation errors. The pipeline comprises three stages: (i) sliding-window aggregation of hidden activations, (ii) extraction of spectral features that capture low-rank structure and stability, and (iii) temporal classification using a lightweight recurrent model. This design rests on two principles: spectral statistics provide global, token-robust indicators of representation geometry, and temporal modeling enables early detection by tracking how uncertainty accumulates across layers and decoding steps. Consistent with our RMT-based hypothesis, hallucination/OOD exhibit spectra that move toward noise baselines and lack clear low-rank spikes.

\subsection{Sliding-Window Representation}

We monitor a set of $m$ transformer layers $L = \{\ell_1,\dots,\ell_m\}$, each producing a hidden activation $h^{\ell}_t \in \mathbb{R}^d$ at generation step $t$. Activations from the selected layers are concatenated to form
\[
v_t = [h^{\ell_1}_t \,\Vert \cdots \Vert\, h^{\ell_m}_t] \in \mathbb{R}^{md}.
\]
To capture local temporal context, we stack the most recent $N$ token representations into a sliding-window matrix
\[
H_t = [v_{t-N+1}, \dots, v_t]^\top \in \mathbb{R}^{N \times md},
\]
which is updated at each decoding step. This construction aggregates cross-layer and short-horizon temporal information while remaining compatible with streaming inference.

\subsection{Spectral Feature Extraction}

Rather than forming the covariance matrix $C_t = \tfrac{1}{N} H_t^\top H_t$, we compute a truncated singular value decomposition
\[
H_t = U_t \Sigma_t V_t^\top,
\]
and recover the corresponding covariance eigenvalues as $\lambda_{t,i} = \sigma_{t,i}^2 / N$. From these eigenvalues we construct a $k$-dimensional spectral feature vector $F_t$ (default $k=22$ discussed in the interpretability section) capturing signal strength and noise alignment, including: (i) leading eigenvalues, (ii) spectral gaps such as $\lambda_{t,1} / \lambda_{t,2}$, (iii) spectral entropy
\[
S_t = -\sum_i p_{t,i}\log p_{t,i}, \quad p_{t,i} = \lambda_{t,i} / \sum_j \lambda_{t,j},
\]
(iv) spectral variance and (v) divergence from the MP reference law via KL divergence and Wasserstein distance. These statistics summarize low-rank structure, dispersion, and proximity to noise-dominated regimes, providing a compact and interpretable representation of activation dynamics.

\subsection{Recurrent Classification}

The sequence of spectral feature vectors $(F_1,\dots,F_T)$ is treated as a multivariate time series and processed by a lightweight recurrent classifier, instantiated as an RNN, GRU, or LSTM. At each step, $F_t$ is input to the recurrent cell, whose hidden state propagates temporal context across decoding steps. A feed-forward output head produces a binary logit corresponding to factual or in-distribution versus hallucinated or anomalous behavior. Because weights are shared across time, the number of parameters is independent of sequence length, enabling efficient modeling of spectral trajectories rather than static snapshots.

\subsection{Hyperparameters and Extensions}

EigenTrack exposes several tunable parameters that control the accuracy-latency tradeoff, including the monitored layer set $L$, sliding-window length $N$, number of spectral features $k$, and recurrent hidden size. The framework naturally extends to multimodal architectures by constructing $H_t$ from cross-modal fusion layers or vision encoder blocks, enabling consistent treatment of LLMs and VLMs. Prior analyses indicate that later transformer layers tend to be more task-specific and informative for downstream probing \cite{vanaken2019answer}, while combining representations from multiple layers can further improve robustness \cite{hosseini2023bert}. In our experiments, the final layers are most predictive when monitored alone, but incorporating earlier layers improves stability across prompts and domains. Accordingly, EigenTrack monitors all layers when feasible, or samples layers at fixed intervals from early to late blocks, which is our default configuration.

\subsection{Computational Overhead}

Let $D = md$ denote the concatenated feature dimension and $N$ the sliding-window length. Computing the SVD of an $N \times D$ matrix requires $O(ND\min\{N,D\})$ operations. In practice, $D > N$, so the dominant cost is $O(N^2 D)$. The per-window computation therefore scales quadratically with $N$ and linearly with the number of monitored layers $m$ and hidden dimension $d$. For a fixed response length, smaller windows yield more updates and higher cost, while larger windows reduce the number of SVDs at the expense of responsiveness. An ablation study in Sec.~\ref{sec:ablation} quantifies this tradeoff and reports wall-clock latency in milliseconds across window sizes.

Additionally, EigenTrack processes activations in a streaming manner and retains only the compact spectral feature vector for each window, rather than storing full activation histories or complete eigenspectra. As a result, memory usage scales linearly with the number of windows and is independent of model size beyond the selected feature dimension.

\begin{figure}[t]
    \centering
    \includegraphics[width=0.5\textwidth]{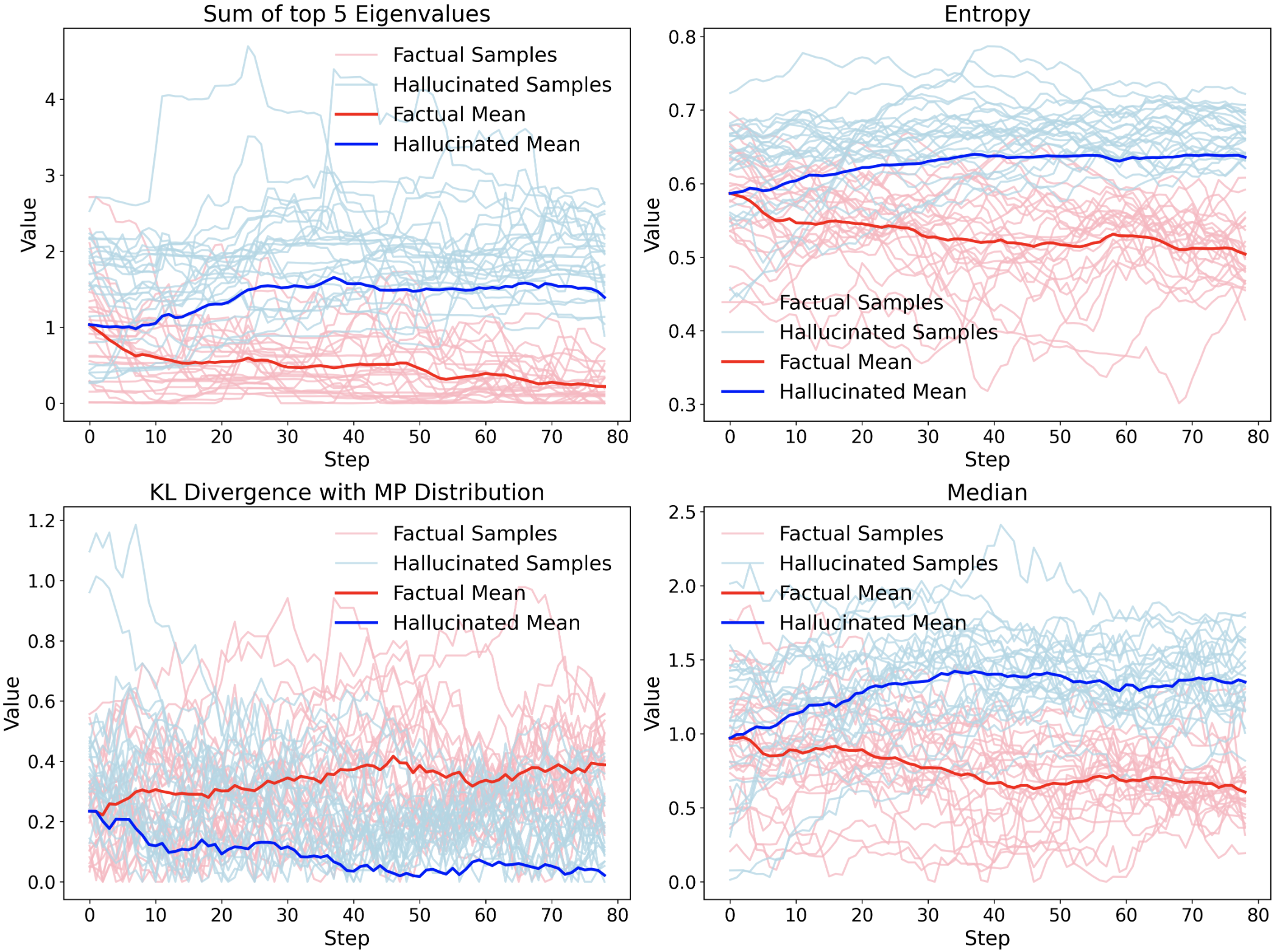}
    \caption{Temporal evolution of spectral features over 80 generation steps on LLaMa-3B for factual and hallucinated sequences.\vspace{-5pt}}
    \label{fig:feature_sequence}
\end{figure}

\subsection{Spectral Dynamics}

Fig.~\ref{fig:feature_sequence} illustrates the temporal behavior of representative spectral features for factual and hallucinated generations. Clear and consistent patterns emerge. Hallucinated sequences concentrate variance along a small number of dominant directions, reflected in larger sums of leading eigenvalues. At the same time, their spectra are flatter and more dispersed, leading to higher entropy. From an RMT perspective, purely unstructured activations would follow the Marchenko-Pastur distribution \cite{marchenko1967distribution}. Relative to this baseline, hallucinated sequences remain closer to the noise regime, exhibiting lower KL divergence and weaker spectral separation, while factual sequences diverge more strongly, indicating structured representation dynamics. The median eigenvalue is also higher and more stable for hallucinations, whereas factual generations show lower and more variable medians that track model confidence. These spectral signatures emerge early in generation and provide reliable indicators of hallucination and OOD behavior before errors become visible in the output.

\begin{table*}[t]
\centering
\normalfont
\setlength{\tabcolsep}{8pt}
\renewcommand{\arraystretch}{1}
\caption{Full results of hallucination and out-of-distribution (OOD) detection across LLMs and VLMs of varying sizes using different recurrent classifiers (RNN, GRU, LSTM). We report AUROC and F1 scores obtained with the full spectral feature set, as well as AUROC values for the best and worst triplets of spectral features identified through ablation.}
\label{tab:auroc_results}
\begin{tabularx}{\textwidth}{c l *{12}{c}}
\cmidrule(lr){3-14}
& & \multicolumn{3}{c}{\textbf{AUROC Full}} & \multicolumn{3}{c}{\textbf{F1 Full}} & \multicolumn{3}{c}{\textbf{AUROC Best Triplet}} & \multicolumn{3}{c}{\textbf{AUROC Worst Triplet}} \\
\cmidrule(lr){3-5}\cmidrule(lr){6-8}\cmidrule(lr){9-11}\cmidrule(lr){12-14}
& \textbf{Model} & RNN & GRU & LSTM & RNN & GRU & LSTM & RNN & GRU & LSTM & RNN & GRU & LSTM \\
\midrule
\multirow{7}{*}{\rotatebox[origin=c]{90}{\textbf{Hallucination}}}
& LLaMa 1B   & 0.799 & 0.842 & 0.831 & 0.750 & 0.790 & 0.783 & 0.725 & 0.774 & 0.758 & 0.597 & 0.644 & 0.630 \\
& LLaMa 3B   & 0.832 & 0.861 & 0.844 & 0.779 & 0.808 & 0.794 & 0.760 & 0.779 & 0.772 & 0.632 & 0.662 & 0.645 \\
& LLaMa 7B   & 0.853 & 0.894 & 0.872 & 0.805 & 0.851 & 0.825 & 0.771 & 0.789 & 0.778 & 0.656 & 0.698 & 0.668 \\
& Qwen 1.8B  & 0.724 & 0.824 & 0.821 & 0.672 & 0.798 & 0.774 & 0.657 & 0.780 & 0.750 & 0.525 & 0.649 & 0.622 \\
& Qwen 7B    & 0.842 & 0.931 & 0.922 & 0.794 & 0.881 & 0.870 & 0.773 & 0.813 & 0.810 & 0.646 & 0.726 & 0.718 \\
& Mistral 7B & 0.864 & 0.888 & 0.871 & 0.812 & 0.839 & 0.819 & 0.791 & 0.817 & 0.799 & 0.662 & 0.687 & 0.675 \\
& LLaVa 7B   & 0.902 & 0.941 & 0.934 & 0.853 & 0.892 & 0.887 & 0.813 & 0.828 & 0.815 & 0.699 & 0.740 & 0.735 \\
\midrule
\multirow{7}{*}{\rotatebox[origin=c]{90}{\textbf{OOD}}}
& LLaMa 1B   & 0.825 & 0.855 & 0.852 & 0.776 & 0.814 & 0.802 & 0.753 & 0.793 & 0.781 & 0.626 & 0.666 & 0.652 \\
& LLaMa 3B   & 0.858 & 0.892 & 0.871 & 0.810 & 0.841 & 0.821 & 0.787 & 0.788 & 0.779 & 0.660 & 0.693 & 0.671 \\
& LLaMa 7B   & 0.879 & 0.924 & 0.897 & 0.829 & 0.874 & 0.847 & 0.801 & 0.807 & 0.805 & 0.680 & 0.724 & 0.692 \\
& Qwen 1.8B  & 0.762 & 0.872 & 0.846 & 0.713 & 0.821 & 0.796 & 0.690 & 0.800 & 0.775 & 0.563 & 0.673 & 0.647 \\
& Qwen 7B    & 0.867 & 0.948 & 0.936 & 0.817 & 0.898 & 0.885 & 0.796 & 0.815 & 0.814 & 0.669 & 0.747 & 0.736 \\
& Mistral 7B & 0.883 & 0.906 & 0.892 & 0.832 & 0.855 & 0.842 & 0.788 & 0.791 & 0.790 & 0.683 & 0.707 & 0.692 \\
& LLaVa 7B   & 0.923 & 0.958 & 0.946 & 0.873 & 0.906 & 0.897 & 0.822 & 0.826 & 0.814 & 0.724 & 0.757 & 0.746 \\
\bottomrule\vspace{-5pt}
\end{tabularx}
\end{table*}

\section{Experimental Setup}

\noindent\textbf{Models and Inference Protocol:}
We evaluate EigenTrack on open-source HuggingFace LLMs and VLMs ranging from 1B to 7B parameters, including models from the LLaMa, Qwen, Mistral, and LLaVa families. Both base and instruction-tuned variants are considered to assess robustness across training regimes. Each model generates up to 128 tokens per prompt, and hidden activations are streamed from selected layers.

\vspace{3pt}
\noindent\textbf{Hallucination Detection:}
Hallucination detection is evaluated on HaluEval, a HotpotQA-based benchmark designed to distinguish factual from hallucinated responses. For each passage, we construct two queries: the ground-truth question associated with the passage, and an unrelated question generated by LLaMa-8B. Responses to the former are treated as factual, while responses to the latter are labeled as hallucinated, following an LLM-as-a-Judge approach. This procedure defines an automatic data-generation pipeline involving three LLMs: (i) a \textit{Main Model}, corresponding to the evaluated model, (ii) a \textit{Question Generator}, which produces unrelated questions, and (iii) an \textit{Answer Judge}, implemented using a larger LLM (LLaMa-8B) that verifies answer consistency. This setup enables scalable generation of labeled hallucinated and non-hallucinated responses without manual annotation. For VLMs such as LLaVa, textual passages are augmented with images from Flickr8k to construct multimodal inputs.

\vspace{3pt}
\noindent\textbf{Out-of-Distribution Detection:}
For OOD detection, we treat WebQuestions as the in-distribution dataset and Eurlex, a legal-domain question dataset, as the OOD set. Eurlex queries are not encountered during pretraining of the evaluated models and therefore induce a clear domain shift. For VLM experiments, the same questions are paired with contextual images from Flickr8k to maintain a consistent multimodal setting.

\vspace{3pt}
\noindent\textbf{EigenTrack Configuration:}
EigenTrack converts streamed hidden activations into sequences of spectral feature vectors, later processed by recurrent classifiers. We evaluate three recurrent architectures: RNN, GRU, and LSTM. Each classifier consists of a linear projection, a single recurrent layer, and a binary output head for in-distribution or factual versus anomalous behavior.

\vspace{3pt}
\noindent\textbf{Baselines and Evaluation Metrics:}
We compare EigenTrack against standard confidence-based and distance-based baselines, including Max Softmax Probability, Energy Score, and cosine-distance variants, as well as recent state-of-the-art methods summarized in Table~\ref{tab:sota_comparison}. All methods are evaluated using the Area Under the Receiver Operating Characteristic curve (AUROC), which ranges from 0.5 for random guessing to 1.0 for perfect discrimination. To ensure fair comparison, all methods use identical train and test splits and the same preprocessing. For score-based baselines, AUROC is computed by sweeping thresholds over in-distribution scores without supervision. In contrast, EigenTrack trains the recurrent classifier in a supervised manner using labeled data. All hyperparameters, including sliding-window length, spectral feature dimensionality, and recurrent hidden size, are selected on a validation set and fixed prior to testing to prevent leakage.

\section{Results}

We evaluate EigenTrack using RNN, GRU, and LSTM recurrent heads and report all results. Across all settings, GRUs consistently achieve the strongest performance. Unless stated otherwise, reported EigenTrack results therefore refer to the GRU variant. For the same reason, ablation studies primarily use GRUs, while differences across recurrent architectures are analyzed explicitly in the interpretability section.

\vspace{3pt}
\noindent\textbf{Hallucination Detection:}
Table~\ref{tab:auroc_results} shows that EigenTrack achieves strong hallucination detection performance across all evaluated LLMs, with AUROC ranging from 0.82 to 0.94. GRUs consistently outperform LSTMs and standard RNNs, highlighting the importance of gated temporal memory for tracking spectral dynamics over decoding steps. Performance improves with model scale: moving from LLaMa-1B to LLaMa-7B yields an average gain of approximately 0.05 AUROC. The strongest results are on 7B-scale models, with Qwen-7B and LLaVa-7B exceeding 0.93 AUROC.
Larger models produce sharper and more stable spectral signatures, which GRUs exploit more effectively than simpler recurrent cells. Compared to prior hallucination detectors evaluated on HaluEval, EigenTrack achieves a clear margin. On LLaMa-7B, it reaches 0.89 AUROC, surpassing HaloScope (0.86), LapEigvals (0.87), INSIDE (0.81), and even SelfCheckGPT applied to the much larger LLaMa-30B (0.84). These results indicate that temporal spectral modeling can compensate for smaller backbone size by extracting more informative internal dynamics.

\vspace{3pt}
\noindent\textbf{OOD Detection:}
A similar pattern emerges for OOD detection, as shown in Table~\ref{tab:auroc_results}. EigenTrack achieves AUROC values between 0.85 and 0.96 across models and architectures. GRU performance improves from 0.855 on LLaMa-1B to 0.924 on LLaMa-7B, and reaches 0.948 on Qwen-7B and 0.958 on LLaVa-7B. The gap between GRUs and RNNs or LSTMs is larger for smaller models, suggesting that gated memory is particularly valuable when spectral signals are weaker or noisier. Across all models, OOD detection performance is typically higher than hallucination detection. This suggests that domain shifts induce more pronounced and consistent spectral deviations than factual drift within the same domain. By combining global spectral features with temporal modeling, EigenTrack achieves state-of-the-art performance on both hallucination and OOD detection while relying only on a lightweight recurrent head.

\vspace{3pt}
\noindent\textbf{Best vs.\ Worst Feature Triplets:}
Table~\ref{tab:auroc_results} also reports the best and worst performing triplets of spectral features identified through ablation. The large gap between these extremes across model sizes and recurrent architectures indicates that detection performance is driven by a compact subset of informative statistics rather than diffuse contributions. GRUs benefit most from well-chosen triplets, while RNNs degrade more sharply under suboptimal feature selection, consistent with their weaker ability to integrate temporal context.

Larger models exhibit smaller gaps between best and worst triplets, suggesting that their activation spectra provide more stable signals even when the feature subset is not optimal. Consistent with the RMT grounding, triplets that include divergence to the Marchenko-Pastur baseline, leading eigenvalues, and spectral gaps are consistently most predictive, whereas triplets dominated by central-tendency measures contribute less. This pattern aligns with the interpretability analysis in Sec.~\ref{sec:interpretability}, which shows that predictive power concentrates on a small set of theoretically motivated spectral features.

\begin{table}[h]
\centering
\caption{AUROC comparison on LLaMa family models (1B/3B/7B) of EigenTrack with GRU head classifier and other SOTA methods.}
\label{tab:sota_comparison}
\small
\setlength{\tabcolsep}{4pt}
\begin{tabular}{c l ccc}
\toprule
& Method & LLaMa-1B & LLaMa-3B & LLaMa-7B \\
\midrule
\multirow{5}{*}{\rotatebox[origin=c]{90}{Hallucination}}
& \textbf{EigenTrack}  & \textbf{0.842} & \textbf{0.861} & \textbf{0.894} \\
& LapEigvals           & 0.785 & 0.819 & \underline{0.871} \\
& INSIDE               & 0.753 & \underline{0.831} & 0.810 \\
& SelfCheckGPT         & 0.739 & 0.804 & 0.809 \\
& HaloScope            & \underline{0.820} & 0.827 & 0.861 \\
\midrule
\multirow{5}{*}{\rotatebox[origin=c]{90}{OOD}}
& \textbf{EigenTrack}  & \textbf{0.855} & \textbf{0.892} & \textbf{0.924} \\
& Cosine Distance      & 0.819 & \underline{0.877} & 0.920 \\
& Energy Score         & \underline{83.2} & 0.852 & 0.890 \\
& Max Softmax Prob     & 0.701 & 0.710 & 0.720 \\
& ODIN                 & 0.801 & 0.842 & \underline{0.921} \\
\bottomrule
\end{tabular}
\end{table}

\vspace{3pt}
\noindent\textbf{Comparison to State of the Art:}
Table~\ref{tab:sota_comparison} compares EigenTrack to representative state-of-the-art detectors on the LLaMa model family. Across both hallucination and OOD settings, EigenTrack consistently achieves the highest AUROC scores. These gains stem from modeling temporal spectral statistics of hidden activations, which capture global representation dynamics that surface-level confidence methods such as Max Softmax Probability and ODIN, as well as snapshot spectral analyses such as LapEigvals, fail to capture. Baseline OOD methods are score-based and operate without OOD supervision. Their AUROC values are obtained by sweeping thresholds over in-distribution scores, ensuring a fair and threshold-independent comparison. Overall, GRUs outperform LSTMs and RNNs, confirming the value of gated memory for modeling spectral evolution, with a lightweight architecture.

EigenTrack outperforms simpler baselines because it jointly models multi-layer and temporal dynamics. RMT-based distances, such as KL divergence and Wasserstein distance to the Marchenko-Pastur baseline, provide a principled reference for quantifying deviation from isotropic noise. The recurrent architecture further captures sequential patterns that static metrics and single-step distances miss, enabling early detection when spectra drift toward noise-like regimes. We also observe that VLMs generally exhibit stronger detection performance than text-only LLMs. A plausible explanation is that cross-modal alignment and redundant visual cues reduce ambiguity, making distribution shifts more detectable in the joint representation. A deeper multimodal analysis is left to future work.

\vspace{3pt}
\noindent\textbf{Interpretability Analysis:}
\label{sec:interpretability}
An important question is which spectral statistics are most critical for detection performance. To address this, we train GRU classifiers on all triplets of spectral features and report results in Table~\ref{tab:auroc_results}. Discriminative power is concentrated in a small subset of features rather than distributed uniformly. In particular, spectral power measures, leading eigenvalues, KL divergence from the Marchenko-Pastur baseline, and intermediate spectral gaps consistently appear in the most informative combinations, yielding AUROC values up to 0.79 on LLaMa-7B. Excluding these features reduces performance to approximately 0.69, indicating that a compact and interpretable feature set is sufficient.

To further validate feature importance, we apply SHAP to trained recurrent classifiers. SHAP assigns contribution values to each feature, enabling fine-grained attribution of model decisions. Fig.~\ref{fig:SHAP} shows the five most influential features per architecture. GRUs emphasize KL divergence, leading eigenvalue gaps, and entropy, reflecting sensitivity to structural instabilities. LSTMs place more weight on top-$k$ eigenvalues and entropy, consistent with their ability to model gradual drifts. Standard RNNs rely more heavily on central-tendency measures alongside entropy and the maximum eigenvalue, consistent with their limited temporal memory.

\begin{figure}[t]
    \centering
    \includegraphics[width=\linewidth]{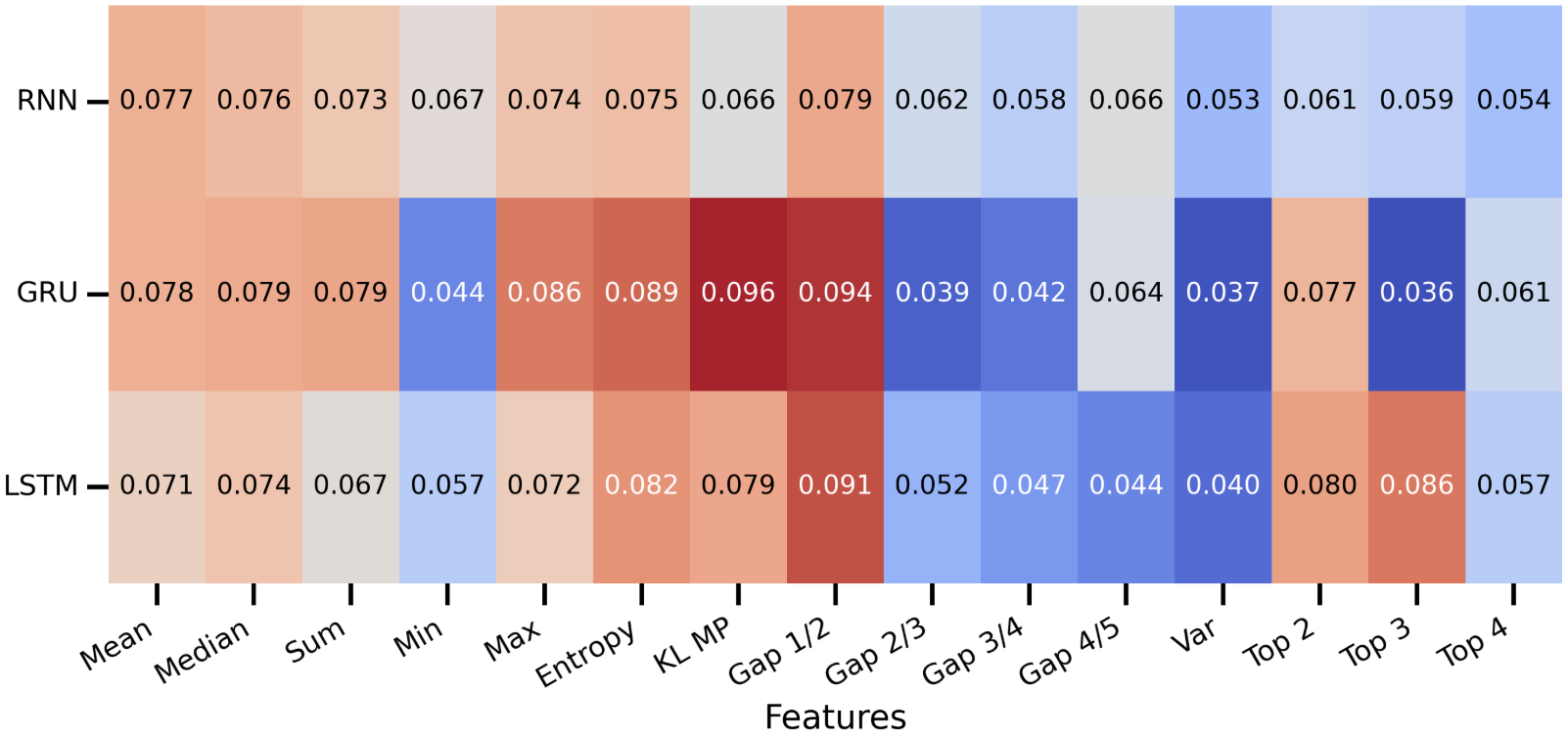}
    \caption{Heatmap (red: high, blue: low) shows the most important features for each classifier on LLaMa 3B and Hallucination dataset, computed by SHAP and normalized. It highlights how different RNNs focus on spectral statistics.\vspace{-5pt}}
    \label{fig:SHAP}
\end{figure}


\begin{figure*}[t]
    \centering
    \includegraphics[width=0.32\linewidth]{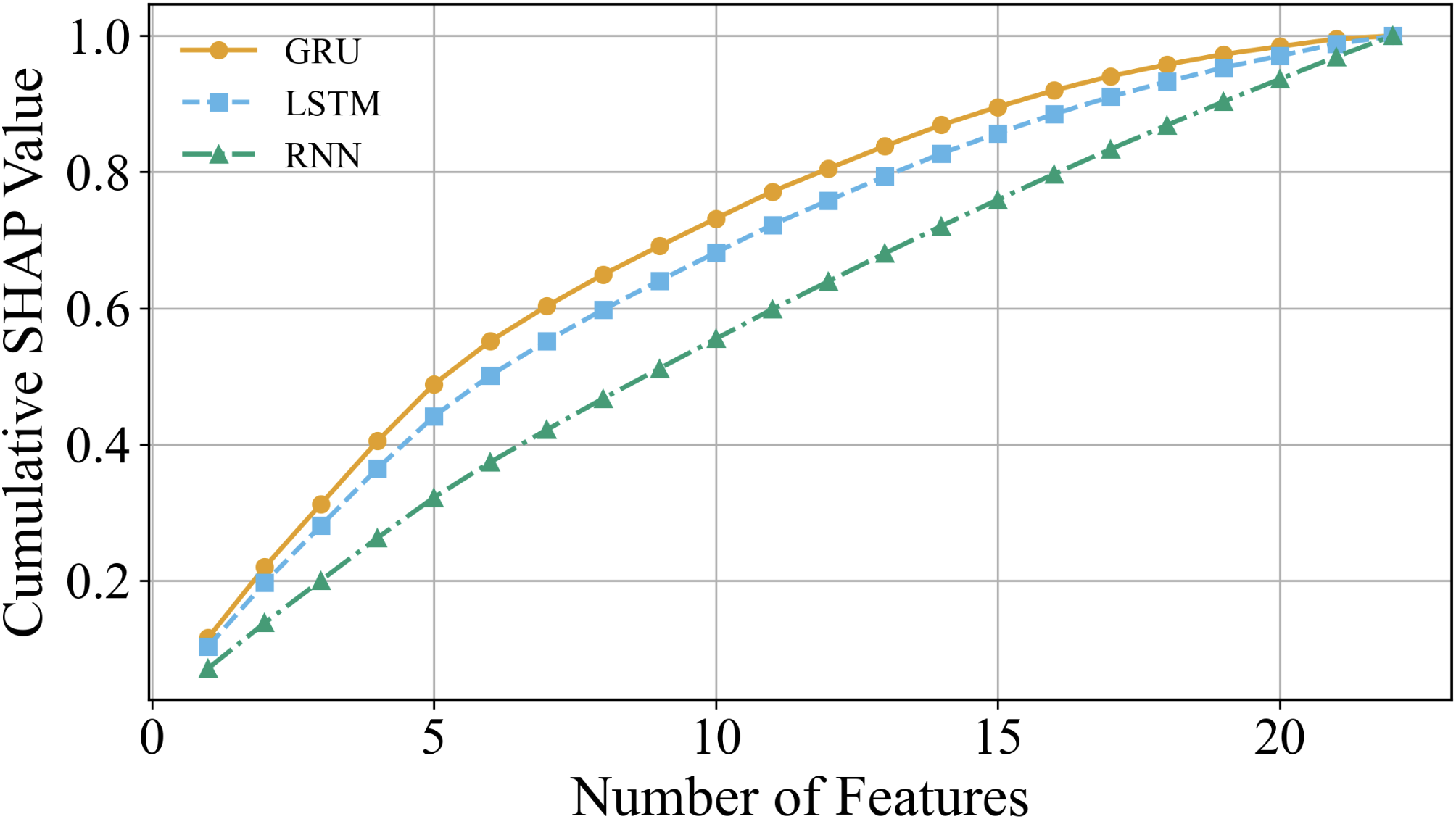}
    \includegraphics[width=0.33\linewidth]
{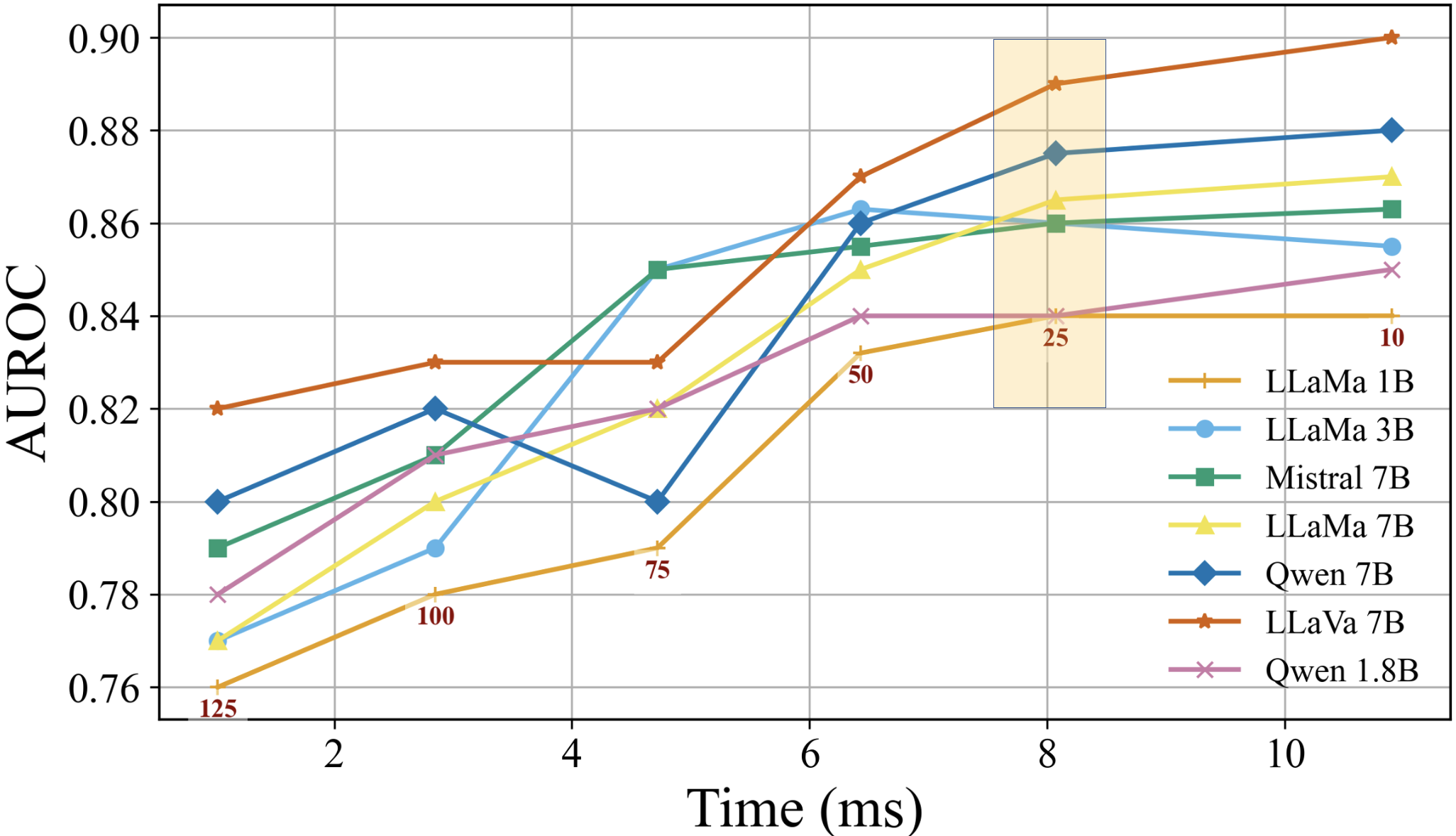}
    \includegraphics[width=0.33\linewidth]{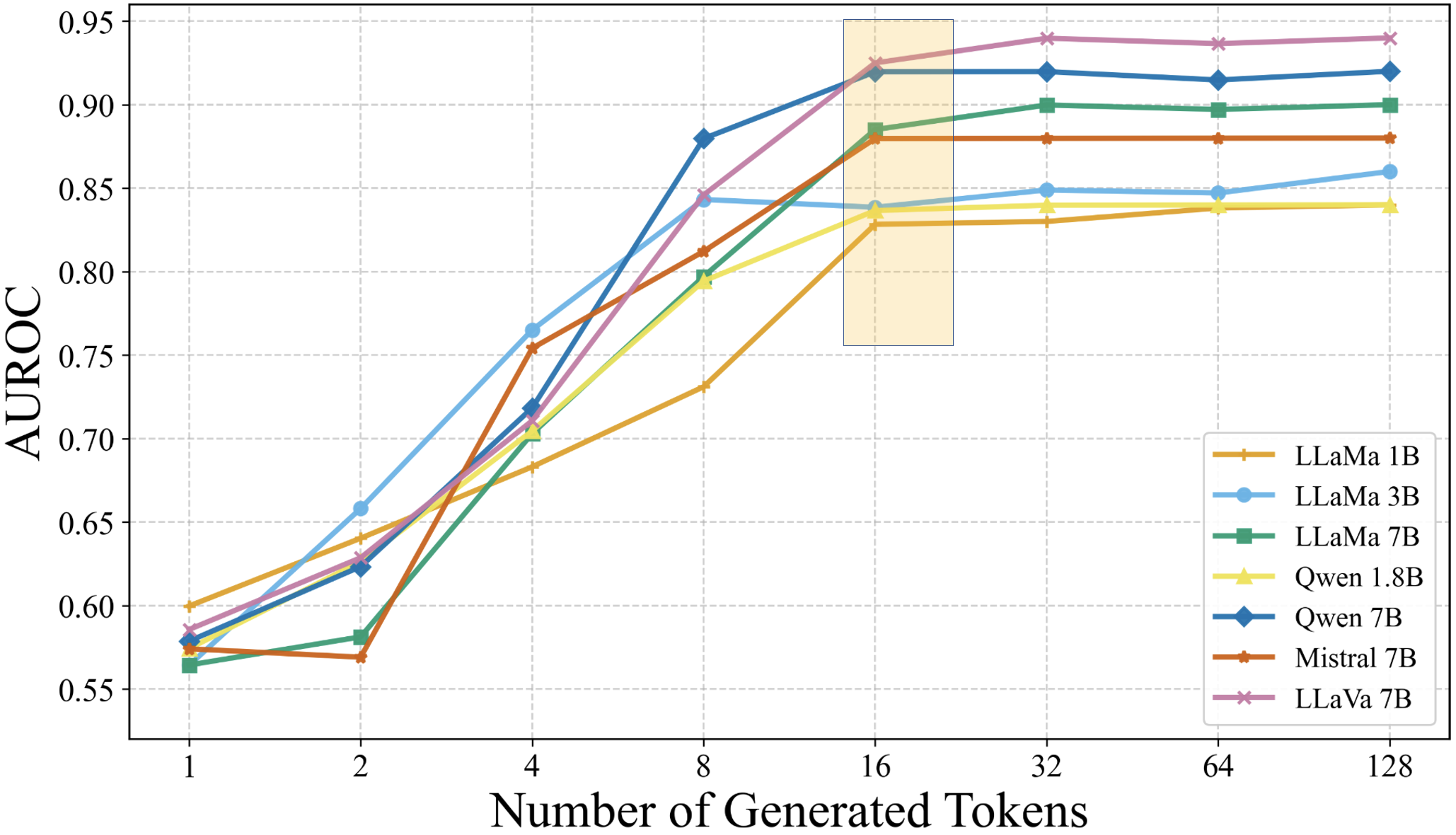}
\caption{(a) Cumulative SHAP attribution mass as a function of the number of top-ranked spectral features for different recurrent architectures on LLaMa-3B (Hallucination dataset). (b) AUROC and inference latency versus sliding-window length for GRU-based hallucination detection on LLaMa-3B (EigenTrack head latency, independent of LLM inference speed). (c) AUROC versus observed response length for GRU-based hallucination detection on LLaMa-3B.\vspace{-5pt}}
    \label{fig:auroc_shap_combined}
\end{figure*}

Fig.~\ref{fig:auroc_shap_combined}(a) summarizes the cumulative SHAP attribution mass as features are added in rank order. GRUs concentrate nearly 80\% of total attribution within the top ten features, while RNNs distribute importance more evenly. LSTMs lie between these extremes, both in feature concentration and overall performance (Table~\ref{tab:auroc_results}). These results confirm that EigenTrack relies on a small set of theoretically grounded spectral descriptors and that gated temporal modeling enables effective exploitation of this structure.

\section{Ablation Studies}
\label{sec:ablation}

We analyze two key hyperparameters that govern the accuracy-latency tradeoff of EigenTrack. The first is the sliding-window length, which determines how much temporal context is aggregated into each spectral snapshot. Short windows respond quickly but are more sensitive to noise, while longer windows smooth fluctuations at the cost of delayed detection. The second is the response length, which controls how many generated tokens are observed before making a decision. Early decisions reduce latency but may miss late-emerging cues, whereas longer responses improve reliability at higher cost. All ablation studies are conducted using LLaMa-3B with a GRU classifier on the hallucination detection task, which serves as a representative and challenging setting.

\subsection{Sliding-Window Length}

Fig.~\ref{fig:auroc_shap_combined}(b) reports AUROC and inference latency as functions of the sliding-window length. Shorter windows capture finer-grained temporal dynamics and yield higher AUROC, but incur greater latency because more windows must be processed per sequence. As the window length increases, performance slowly declines while latency decreases.

A clear knee emerges in the range of 25 to 50 tokens, which provides a strong operating point across models. Below this range, spectral estimates become noisy, and performance stabilizes while computational cost increases. Beyond it, additional temporal context degrades and delays detection capabilities, adding unnecessary smoothing in the sequence. This behavior indicates that the spectral dynamics relevant for hallucination detection are captured within a moderate temporal horizon.

\subsection{Response Length}

Fig.~\ref{fig:auroc_shap_combined}(c) plots AUROC as a function of the number of generated tokens observed before classification. Performance starts near chance and rises sharply within the first few tokens before saturating. This indicates that spectral cues associated with hallucination emerge early in generation, well before errors become explicit in the output. The rapid AUROC increase is consistent with uncertainty accumulating quickly as representations drift toward noise-like regimes, as predicted by the RMT-based analysis, while the plateau reflects diminishing marginal information from later tokens. These results support EigenTrack’s suitability for early stopping, allowing practitioners to trade accuracy for latency by adjusting response length and terminating failing generations without full output.

\section{Conclusion}
We presented EigenTrack, a lightweight and interpretable framework for detecting hallucination and out-of-distribution behavior in LLMs and VLMs via temporal modeling of spectral features. By combining global eigenvalue-based descriptors with RMT-grounded divergence to the Marchenko-Pastur baseline and recurrent temporal modeling, EigenTrack captures representation dynamics that static or output-layer heuristics miss, enabling early detection and efficient termination. Experiments demonstrate state-of-the-art performance across LLM and VLM families with favorable accuracy-latency tradeoffs. Future work will focus on tighter theoretical guarantees, deeper multimodal analysis, adaptive correction strategies, and automatic selection of spectral features, window size, and monitored layers.

\bibliographystyle{IEEEtran}
\bibliography{refs}

@article{ettori2025rmtkdrandommatrixtheoretic,
  title   = {RMT-KD: Random Matrix Theoretic Causal Knowledge Distillation},
  author  = {Davide Ettori and Nastaran Darabi and Sureshkumar Senthilkumar and Amit Ranjan Trivedi},
  journal = {arXiv preprint arXiv:2509.15724},
  year    = {2025}
}

@inproceedings{song2022rankfeat,
  title     = {RankFeat: Rank-1 Feature Removal for Out-of-Distribution Detection},
  author    = {Song, Yixin and Huang, Yue and Yang, Chenglu and Shi, Yicheng and Wei, Dong and Sun, Lei and Chen, Lin},
  booktitle = {Advances in Neural Information Processing Systems (NeurIPS)},
  year      = {2022}
}

@article{mei2024snojoe,
  title   = {Spectral Normalized Joint Energy for Multi-Label Out-of-Distribution Detection},
  author  = {Mei, Yihan and Li, Zhiyu and Li, Yiyou and Zou, James},
  journal = {arXiv preprint arXiv:2405.04759},
  year    = {2024}
}

@article{gu2025spectralgap,
  title   = {SpectralGap: Graph-Level Out-of-Distribution Detection via Laplacian Eigenvalue Gaps},
  author  = {Gu, Jiaqi and Qiao, Yibo and Li, Pan},
  journal = {arXiv preprint arXiv:2505.15177},
  year    = {2025}
}

@article{darabi2025eigenshield,
  title={EigenShield: Causal Subspace Filtering via Random Matrix Theory for Adversarially Robust Vision-Language Models},
  author={Darabi, Nastaran and Naik, Devashri and Tayebati, Sina and Jayasuriya, Dinithi and Krishnan, Ranganath and Trivedi, Amit Ranjan},
  journal={arXiv preprint arXiv:2502.14976},
  year={2025}
}

@inproceedings{tayebati2025cap,
  author    = {Sina Tayebati and Nastaran Darabi and Divake Kumar and 
               Dinithi Kasthuriarachchi Kankanamalage Dona and 
               Theja Tulabandhula and Ranganath Krishnan and 
               Amit Ranjan Trivedi},
  title     = {CAP: Conformalized Abstention Policies for Context-Adaptive Risk Management for LLMs and VLMs},
  booktitle = {Proceedings of the Asian Conference on Machine Learning (ACML)},
  year      = {2025}
}

@inproceedings{jayasuriya2025sparc,
  author    = {Dinithi Jayasuriya and Sina Tayebati and Davide Ettori and Ranganath Krishnan and Amit Ranjan Trivedi},
  title     = {SPARC: Subspace-Aware Prompt Adaptation for Robust Continual Learning in LLMs},
  booktitle = {Proceedings of the 2025 International Joint Conference on Neural Networks},
  year      = {2025}
}

@article{marchenko1967distribution,
  title     = {Distribution of eigenvalues for some sets of random matrices},
  author    = {Mar{\v{c}}enko, V. A. and Pastur, L. A.},
  journal   = {Mathematics of the USSR-Sbornik},
  year      = {1967},
  publisher = {IOP Publishing},
  doi       = {10.1070/SM1967v001n04ABEH001994}
}

@article{radford2019language,
  title={Language models are unsupervised multitask learners},
  author={Radford, Alec and Wu, Jeffrey and Child, Rewon and Luan, David and Amodei, Dario and Sutskever, Ilya},
  journal={OpenAI Blog},
  volume={1},
  number={8},
  pages={9},
  year={2019}
}

@article{baik2005phase,
  title={Phase transition of the largest eigenvalue for nonnull complex sample covariance matrices},
  author={Baik, Jinho and Ben Arous, Gérard and Péché, Sandrine},
  journal={Annals of Probability},
  volume={33},
  number={5},
  pages={1643--1697},
  year={2005},
  publisher={Institute of Mathematical Statistics}
}

@article{kobayashi2024weight,
  title={Weight decay induces low-rank attention layers},
  author={Kobayashi, Seijin and Akram, Yassir and von Oswald, Johannes},
  journal={arXiv preprint arXiv:2410.23819},
  year={2024}
}

@article{farquhar2024entropy,
  author  = {Sebastian Farquhar and Yarin Gal},
  title   = {Semantic Entropy: Language Models Detect and Explain their Failures},
  journal = {Transactions on Machine Learning Research},
  year    = {2024},
  note    = {arXiv:2305.17409}
}

@inproceedings{sriramanan2024attention,
  author    = {Anand Sriramanan and Sheng Shen and Tri Dao and others},
  title     = {AttentionScore: Faithfulness Detection from Attention Patterns in LLMs},
  booktitle = {Proceedings of ICML 2024},
  year      = {2024},
  note      = {arXiv:2311.09516}
}

@inproceedings{hendrycks2017baseline,
  title     = {A Baseline for Detecting Misclassified and Out-of-Distribution Examples in Neural Networks},
  author    = {Hendrycks, Dan and Gimpel, Kevin},
  booktitle = {International Conference on Learning Representations},
  year      = {2017}
}

@inproceedings{su2024mind,
  title     = {Unsupervised Real-Time Hallucination Detection Based on the Internal States of Large Language Models},
  author    = {Su, Weihang and Wang, Changyue and Ai, Qingyao and Hu, Yiran and Wu, Zhijing and Zhou, Yujia and Liu, Yiqun},
  booktitle = {Findings of the Association for Computational Linguistics: ACL 2024},
  pages     = {14379--14391},
  year      = {2024}
}

@inproceedings{binkowski2025lapeigvals,
  title     = {Hallucination Detection in LLMs Using Spectral Features of Attention Maps},
  author    = {Binkowski, Jakub and Janiak, Denis and Sawczyn, Albert and Gabrys, Bogdan and Kajdanowicz, Tomasz},
  booktitle = {Proceedings of ICML 2025},
  year      = {2025}
}

@inproceedings{sun2025redeep,
  title     = {ReDeEP: Detecting Hallucination in Retrieval-Augmented Generation via Mechanistic Interpretability},
  author    = {Sun, Zhongxiang and Zang, Xiaoxue and Zheng, Kai and Song, Yang and Xu, Jun and Zhang, Xiao and Yu, Weijie and Li, Han},
  booktitle = {Proceedings of ACL 2025},
  year      = {2025}
}

@inproceedings{selfcheckgpt,
  title     = {SelfCheckGPT: Zero-Resource Black-Box Hallucination Detection for Generative Large Language Models},
  author    = {Manakul, Potsawee and Liusie, Adian and Gales, Mark J. F.},
  booktitle = {Proceedings of EMNLP 2023},
  pages     = {9004--9017},
  year      = {2023}
}

@article{conli,
  title   = {Chain of Natural Language Inference for Reducing Large Language Model Ungrounded Hallucinations},
  author  = {Lei, Deren and Li, Yaxi and Hu, Mengya and Wang, Mingyu and Yun, Vincent and Ching, Emily and Kamal, Eslam},
  journal = {arXiv preprint arXiv:2310.03951},
  year    = {2023}
}

@article{costeffective,
  title   = {Cost-Effective Hallucination Detection for LLMs},
  author  = {Valentin, Simon and Fu, Jinmiao and Detommaso, Gianluca and Xu, Shaoyuan and Zappella, Giovanni and Wang, Bryan},
  journal = {arXiv preprint arXiv:2407.21424},
  year    = {2024}
}

@inproceedings{detectgpt,
  title     = {DetectGPT: Zero-Shot Machine-Generated Text Detection using Probability Curvature},
  author    = {Mitchell, Eric and Lee, Yoonho and Khazatsky, Alexander and Manning, Christopher~D. and Finn, Chelsea},
  booktitle = {Proceedings of ICML 2023},
  pages     = {24950--24962},
  year      = {2023}
}

@inproceedings{glimpse,
  title     = {Glimpse: Enabling White-Box Methods to Use Proprietary Models for Zero-Shot LLM-Generated Text Detection},
  author    = {Bao, Guangsheng and Zhao, Yanbin and He, Juncai and Zhang, Yue},
  booktitle = {Proceedings of ICLR 2025},
  year      = {2025}
}

@article{fastdetectgpt,
  title   = {Fast-DetectGPT: Efficient Zero-Shot Detection of Machine-Generated Text via Conditional Probability Curvature},
  author  = {Bao, Guangsheng and Zhao, Yanbin and Teng, Zhiyang and Yang, Linyi and Zhang, Yue},
  journal = {arXiv preprint arXiv:2310.05130},
  year    = {2023}
}

@article{wigner1958distribution,
  title   = {On the distribution of the roots of certain symmetric matrices},
  author  = {Wigner, Eugene P.},
  journal = {Annals of Mathematics},
  volume  = {67},
  number  = {2},
  pages   = {325--327},
  year    = {1958}
}

@inproceedings{pennington2017rmt,
  title     = {Nonlinear Random Matrix Theory for Deep Learning},
  author    = {Pennington, Jeffrey and Worah, Pratik},
  booktitle = {Advances in Neural Information Processing Systems},
  year      = {2017}
}

@article{martin2017implicit,
  title   = {Implicit Self-Regularization in Deep Neural Networks: Evidence from Random Matrix Theory and Implications for Learning},
  author  = {Martin, Charles H. and Mahoney, Michael W.},
  journal = {arXiv preprint arXiv:1710.09553},
  year    = {2017}
}

@inproceedings{hosseini2023bert,
  title     = {BERT Has More to Offer: BERT Layers Combination Yields Better Sentence Embeddings},
  author    = {Hosseini, MohammadSaleh and Munia, Munawara and Khan, Latifur},
  booktitle = {Findings of the Association for Computational Linguistics: EMNLP 2023},
  pages     = {15419--15431},
  year      = {2023},
  address   = {Singapore},
  publisher = {Association for Computational Linguistics}
}

@inproceedings{vanaken2019answer,
  title     = {How Does BERT Answer Questions? A Layer-Wise Analysis of Transformer Representations},
  author    = {van Aken, Betty and Winter, Benjamin and Loeser, Alexander and Gers, Felix A.},
  booktitle = {Proceedings of the 28th ACM International Conference on Information and Knowledge Management (CIKM)},
  year      = {2019},
  doi       = {10.1145/3357384.3358028},
  note      = {arXiv:1909.04925},
  url       = {https://arxiv.org/abs/1909.04925}
}

@article{paul2007asymptotics,
  title   = {Asymptotics of sample eigenstructure for a large dimensional spiked covariance model},
  author  = {Paul, Debashis},
  journal = {Statistica Sinica},
  year    = {2007}
}

@inproceedings{stutts2023lightweight,
  title={Lightweight, uncertainty-aware conformalized visual odometry},
  author={Stutts, Alex C and Erricolo, Danilo and Tulabandhula, Theja and Trivedi, Amit Ranjan},
  booktitle={2023 IEEE/RSJ International Conference on Intelligent Robots and Systems (IROS)},
  pages={7742--7749},
  year={2023},
  organization={IEEE}
}

@inproceedings{stutts2024mutual,
  title={Mutual information-calibrated conformal feature fusion for uncertainty-aware multimodal 3d object detection at the edge},
  author={Stutts, Alex C and Erricolo, Danilo and Ravi, Sathya and Tulabandhula, Theja and Trivedi, Amit Ranjan},
  booktitle={2024 IEEE international conference on robotics and automation (ICRA)},
  pages={2029--2035},
  year={2024},
  organization={IEEE}
}

\end{document}